\documentclass[journal]{IEEEtran}
\usepackage{units}
\usepackage{cite}
\usepackage{array}
\usepackage{subfigure}
\usepackage{graphicx}
\usepackage{amsmath}
\usepackage{amssymb}
\usepackage{mathptmx}
\usepackage{epstopdf}
\graphicspath{}
\usepackage{epsfig}
\usepackage{color}
\usepackage[normalem]{ulem}
\usepackage{wasysym}
\usepackage{fancyhdr}
\usepackage{multirow}
\usepackage{algpseudocode}

\begin{document}

\title{HFirst: A Temporal Approach to Object Recognition}

\author{ {Garrick~Orchard, Cedric~Meyer, Ralph~Etienne-Cummings~\textit{Fellow,~IEEE}, Christoph~Posch~\textit{Senior~Member,~IEEE}, Nitish~Thakor~\textit{Fellow,~IEEE}, Ryad~Benosman}
\thanks{Manuscript received 12 November 2013, revised 19 December 2014, accepted January 7 2015.}
\thanks{Collaboration on this work was supported by the Merlion Programme of the Institut Français de Singapour, under administrative supervision of the French Ministry of Foreign Affairs and the National University of Singapore. The views and conclusions contained in this document are those of the authors and should not be interpreted as representing the official policies, either expressly or implied, of the French Ministry of Foreign Affairs.}
\thanks{This research was funded by the SINAPSE startup grant from the National University of Singapore and Singapore Ministry of Defence.}
\thanks{Garrick Orchard and Nitish Thakor are with the Singapore Institute for Neurotechnology (SINAPSE) at the National University of Singapore, Singapore (Email:~garrickorchard@nus.edu.sg,~eletnv@nus.edu.sg)}
\thanks{Cedric Meyer, Christoph Posch, and Ryad Benosman are with Institut De la Vision at Universite Pierre et Marie Curie, Paris, France (Email: meyer.cece@gmail.com, cposch@yahoo.com, ryad.benosman@upmc.fr)}
\thanks{Ralph Etienne-Cummings is with the department of Electrical and Computer Engineering at Johns Hopkins University, Baltimore MD, USA  (Email:~retienne@jhu.edu)}
}

\maketitle

\begin{abstract}
This paper introduces a spiking hierarchical model for object recognition which utilizes the precise timing information inherently present in the output of biologically inspired asynchronous Address Event Representation (AER) vision sensors.
The asynchronous nature of these systems frees computation and communication from the rigid predetermined timing enforced by system clocks in conventional systems. Freedom from rigid timing constraints opens the possibility of using true timing to our advantage in computation. We show not only how timing can be used in object recognition, but also how it can in fact simplify computation. Specifically, we rely on a simple temporal-winner-take-all rather than more computationally intensive synchronous operations typically used in biologically inspired neural networks for object recognition. This approach to visual computation represents a major paradigm shift from conventional clocked systems and can find application in other sensory modalities and computational tasks. We showcase effectiveness of the approach by achieving the highest reported accuracy to date (97.5\%$\pm$3.5\%) for a previously published four class card pip recognition task and an accuracy of 84.9\%$\pm$1.9\% for a new more difficult 36 class character recognition task.
\end{abstract}

\IEEEpeerreviewmaketitle

\section{Introduction}
This paper tackles the problem of object recognition using a hierarchical Spiking Neural Network (SNN) structure.
We present a model developed for object recognition, which we have called HFirst. The name arises because the approach extensively relies on the first spike received during computation to implement a non-linear pooling operation, which is typically required by frame-based Convolutional Neural Networks (CNNs).

We rely on the biological observation that strongly activated neurons tend to fire first \cite{SpikeTimingGawne01081996, SpikeTimingGreschner01122006}.
In particular, we focus on the relative timing of spikes across neurons, namely the order in which neurons fire. We will argue that such a scheme allows us to derive temporal features that are particulary suited for robust and rapid object recognition at a very low computational cost.
Existing work on artificial neural networks tend to assume a predetermined timing which is completely independent of the processing taking place. This prohibits these artificial NNs from using time in their computation. However, the timing of communication (spikes) in biological networks is known to be very important.
Much like biological networks, in this paper we exploit spike timing to our advantage in computation. More specifically we rely on the time at which a spike is received to implement a simple non-linear operation which replaces the more computationally intensive maximum operation typically used in non-spiking neural networks for visual processing.

Artificial Neural Networks (NNs), of which CNNs are a subset, have successfully been used in many applications, including signal and image processing \cite{masters1993practical, egmont2002image}, and pattern recognition \cite{bishop1995neural}, while hardware acceleration of such models allows real-time operation on megapixel resolution video \cite{NeuFlowISCAS2010}. Although CNN models are argued to be biologically inspired, their artificial implementations are typically far removed from biological neural networks, most of which consist of spiking neurons.

Spiking Neural Networks (SNNs) have received a lot of attention recently as new, more efficient computing technologies are sought as conventional CMOS technology approaches its fundamental limits. { SNNs have the potential to achieve incredibly high power efficiency. This is not a claim that we provide our own evidence for, but is rather based on observations of power consumption in biology (the human brain consumes only 20W) and recent works which present SNNs on chip with impressive power efficiency. Examples include Neurogrid \cite{NeuroGrid} and IBM’s TrueNorth \cite{Merolla2014} which can simulate 1 million spiking neurons while consuming under 100mW. In this paper we address the question of how SNNs can be used for visual object recognition.}

Modern reconfigurable custom SNN hardware platforms can implement hundreds of thousands to millions of spiking neurons in parallel. Examples of these hardware implementation projects include the Integrate and Fire Array Transceiver (IFAT) \cite{IFATGoldberg2001781}, Hierarchical AER-IFAT \cite{HiAER}, Brain Scales \cite{BrainScales}, Spiking Neural Network Architecture (SpiNNaker) \cite{Spinnaker}, Neurogrid \cite{NeuroGrid}, { Qualcomm's Zeroth Processor, and IBM's TrueNorth \cite{Merolla2014} (fabricated with Samsung)}.

In parallel with these hardware platforms, software platforms for neural computation have emerged, including the Neural Engineering Framework (NEF) \cite{NENGO}, Brian \cite{Brian}, and PyNN \cite{PYNN}, many of which can be used to configure the hardware platforms previously mentioned. Continued interest and funding from the European Union's Human Brain Project \cite{HumanBrainProject} and the USA's Brain Research through Advancing Innovative Neurotechnologies (BRAIN) project \cite{BRAINusa} will drive development of such systems for years to come.

As neural simulation hardware matures, so must the algorithms and architectures which can take advantage of this hardware. However, it does not necessarily make sense to directly convert existing computer vision models and algorithms (which process traditional frame based data) to SNN implementation. A central concept within SNNs is that spike timing encodes information, but frames do not contain precise timing information. The timing of the arrival of frames is purely a function of the front end sensor and is completely independent of the scene or stimuli present. In order for a SNN to exploit precise timing, it must operate on data which contains precise timing information and not spike timings artificially generated from frame-based outputs. To obtain visual data with precise timing, we turn our attention to asynchronous AER vision sensors, sometimes referred to as ``silicon retinae" \cite{OctopusRetina, OriginalRetina}. These sensors more closely match the operation of biological retina and do not utilize frames.

Asynchronous AER vision sensors have seen much improvement since their introduction in the early 1990s by Mahowald \cite{mahowald1994analog}. Modern change detection AER sensors reliably provide information on changes of illumination at the focal plane over a wide dynamic range and under a variety of lighting conditions. The pixels within such sensors each contain a circuit which continuously performs local analog computation to detect the occurrence and time of changes in intensity for that particular pixel. This computation at the focal plane is a form of redundancy suppression, ensuring that pixels only output data when new information is present (barring some background noise). Furthermore, the time of arrival of data from the sensor accurately represents when the intensity change occurred. Under test conditions sub-microsecond accuracy is achieved, versus accuracy on the order of milliseconds for fast frame-based cameras. This temporal accuracy provides precise spike timing information which can be exploited by a SNN.

Much like SNNs are a more accurate approximation of biological processing hardware, AER vision sensors are a more accurate approximation of the biological retina. The single bit of data provided by a pixel can be likened to a neural spike, and much like a biological retina, the AER sensor performs computation at the focal plane. Notable examples of spiking AER vision sensors include the earliest examples of spiking silicon retinae by Culurciello~\emph{et~al.} \cite{OctopusRetina} and Zaghloul~\emph{et~al.} \cite{OriginalRetina}, as well as the more recent Dynamic Vision Sensor (DVS) from Delbruck \cite{DVStobi}, the sensitive DVS from Linares-Berranco \cite{BernabeRetina}, and the Asynchronous Time-based Image Sensor (ATIS) from Posch \cite{ATIS}. Operation of these sensors will be discussed in Section \ref{sec:DVS}. For a review of asynchronous event-based vision sensors see Delbruck~\emph{et~al.} \cite{ReviewDelbruck}.

With the emergence of these asynchronous vision sensors, many researchers have taken an interest in processing their data in a manner which takes advantage of the asynchronous, high temporal resolution, and sparse representation of the scene they provide. Models of early visual area V1, including saliency, attention, foveation, and recognition \cite{folowosele2008real, Vogelstein05saliency, vogelstein2007multichip} have been implemented by combining the reconfigurable IFAT system \cite{IFATjacob} with the Octopus silicon retina \cite{OctopusRetina}. More recent focuses in the field include stereo vision \cite{3DAER1, 3DAER2, 3DAER3}, motion estimation \cite{benosman2012asynchronousFLOW, Orchard2014b}, tracking \cite{goalKeeperRetina}, and more object recognition works \cite{BernabeConvNet, CAVIAR, ShouShunAER}. Further information on neuromorphic sensory systems can be found in Liu and Delbruck \cite{reviewNeuromorphicSensorySystems}.

In this paper we focus on the task of object recognition. The most similar recent works include a VLSI implementation of the HMAX model \cite{HMAX2007, Sciences2006} for recognition which uses spiking neurons throughout \cite{FOPEHMAX}. The VLSI spiking HMAX implementation computes all the functions required by HMAX, but operates on 24$\times$24 pixel images, limited by the number of available neurons, and does not run real-time. Adaptations of frame-based CNN techniques for training SNNs and implementing them in FPGA have also been recently presented \cite{shoushun2009bio}, including a recent PAMI paper \cite{BernabeConvNet} which presented a high speed card pip recognition task which we also tackle in this paper as a comparison to existing works.

In this paper we present our SNN architecture dubbed ``HFirst", which takes advantage of timing information provided by AER sensors. A key aspect is that our architecture uses spike timing to encode the strength of neuron activation, with stronger activated neurons spiking earlier. This enables us to implement a MAX operation using a simple temporal Winner-Take-All (WTA) rather than performing a synchronous MAX operation as is typically done in frame-based algorithms \cite{HMAX2007}. Unlike the frame-based MAX operation, which outputs a number representing the strength of the strongest input, the temporal WTA can only output a spike, but by responding with low latency to its inputs, the temporal WTA preserves the time encoding of signal strength. It should be noted that other methods of implementing a MAX operation in spikes have been presented previously \cite{vogelstein2007multichip}.

Masquelier~\emph{et~al.} \cite{masquelier2007unsupervised} also use a temporal WTA, but their approach focuses on static images and spike generation from these images is artificially simulated, whereas we use AER vision sensors \cite{DVStobi, BernabeRetina, ATIS} to directly capture data from dynamic scenes for recognition. Additionally, Masquelier~\emph{et~al.} require their network to be reset before a second object can be recognised, whereas HFirst operates on streaming ``video" and can recognise multiple objects in sequence, or even simultaneously.

The HFirst model described here can be used with many of the available AER change detection sensors, and could be implemented on one of many neural processing platforms. For this particular work we analysed HFirst in simulation using a combination of C and Matlab on a desktop PC. Once simulated, the SNN was implemented in real-time on a Xilinx Spartan 6 XC6SLX150-2 FPGA.

The rest of this paper is organized as follows. In the next section we briefly describe the event-based vision sensors, then we describe the neuron model using spike timing for computation in Section~\ref{sec:ComputingWithNeurons}. The HFirst architecture is described in Section~\ref{sec:AsyncArchitecture}, followed by brief analysis of the required computation and real-time implementation. Testing and results are then presented to showcase the model accuracy before wrapping up with discussions and conclusions.

\section{Asynchronous Change Detection Vision Sensors}
\label{sec:DVS}

Neuromorphic, event-based vision sensors are a novel type of vision sensor driven by changes within the visual scene, much like the human retina, and differs from conventional image sensors which use artificial timing to control information acquisition. The sensors used in this paper \cite{ATIS, DVStobi, BernabeRetina} consist of autonomous pixels, each asynchronously generating spike events that encode relative changes in illumination. These sensors capture visual information at a much higher temporal resolution than conventional vision sensors, achieving accuracy down to sub-microsecond levels under optimal conditions. Moreover, since the pixels only detect temporal changes, temporally redundant information is not captured or communicated, resulting in a sparse representation of the scene. Captured events are transmitted asynchronously by the sensor in the form of continuous-time digital words containing the address of the activated pixel using the AER protocol \cite{mahowald1994analog}.

\begin{figure}
\includegraphics[width=0.5\textwidth]{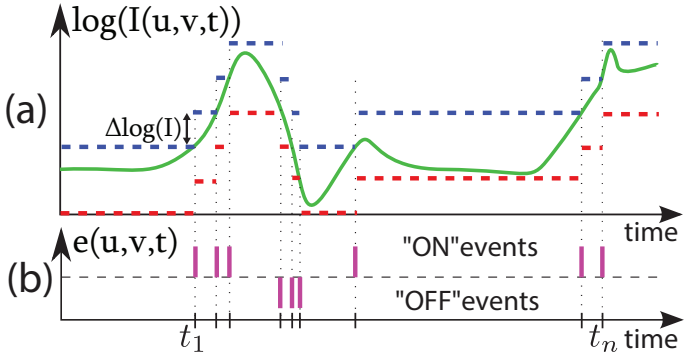}
\caption{Event-based vision sensor acquisition principle. (a) typical signal showing the $log$ of luminance of a pixel located at $[u,v]^T$. Dotted lines show how the thresholds for detecting increases and decreases in intensity change as outputs are generated. (b) asynchronous temporal contrast events generated by this pixel in response to the light variation shown in (a).}
\label{fig:ATISSig}
\end{figure}

To better understand the operation of these sensors we will briefly provide a formulation to approximate the sensor response to visual stimuli. Let us define $I(u,v,t)$ as the intensity of a pixel located at $[u,v]^T$, where $u$ and $v$ are spatial co-ordinates in units of pixels. Each pixel of the sensor asynchronously generates events at the precise time when change in the $log$ of the pixel illumination $\Delta log(I(u,v,t))$ is larger than a certain threshold $\Delta I$ since the last event, as shown Fig. \ref{fig:ATISSig}(a) and (b). The logarithmic relation means the pixels respond to percentage changes in illumination rather than the absolute magnitude of the change. This allows pixels to operate over a very wide dynamic range ($\geq$120dB).

\label{sec:ATIS}
Under constant scene illumination the intensity changes seen by the sensor are due to the combination of a spatial image gradient and a component of image motion along that gradient. As described by the equation below which is a first order approximation of the image constancy constraint.
\begin{equation}\label{eq:ImageConstancy}
\begin{array}{l l}
\large{\frac{\displaystyle dI(u,v,t)}{\displaystyle dt} = -\frac{\displaystyle dI(u,v,t)}{\displaystyle du}\frac{\displaystyle du}{\displaystyle dt} -\frac{\displaystyle dI(u,v,t)}{\displaystyle dv}\frac{\displaystyle dv}{\displaystyle dt}}\\
\end{array}
\end{equation}
{ where $I(u,v,t)$ is intensity on the image plane, and $u$ and $v$ are horizontal and vertical coordinates measured in units of pixels}. The sensor will therefore generate the most events at locations where a large image gradient is present, as will be discussed further in Section~\ref{sec:timing}.

\section{Computing with Neurons}
\label{sec:ComputingWithNeurons}

\begin{figure}
  \centering
  \includegraphics[width=0.5\textwidth]{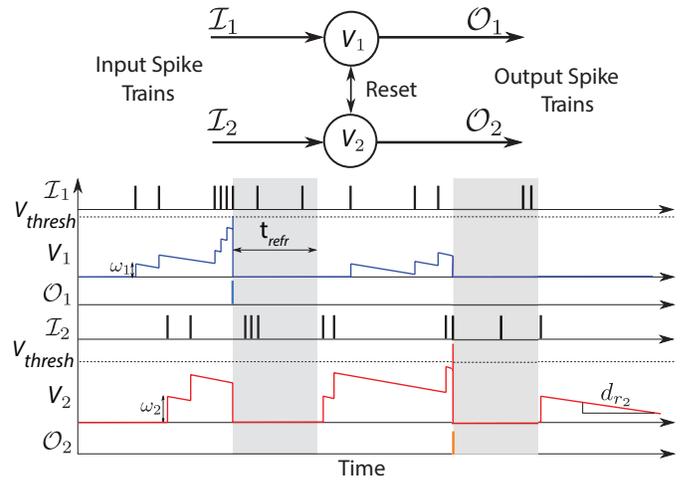}\\
  \caption{Operation of an Integrate-and-Fire neurons (IF neurons) used, showing how synaptic weights and time affect the neuron membrane potential, as well as the operation of {lateral reset connections (lateral meaning connecting to other neurons in the same layer)} and refractory period.}\label{fig:Neuron}
\end{figure}

\subsection{Neuron model}
\label{sec:neuron_model}
The neuron model we use is a simple Integrate-and-Fire neuron (IF neuron) \cite{izhikevich2004model} with linear decay and a refractory period, as shown in Fig.~\ref{fig:Neuron}. We foresee that the model would translate to hardware implementations which model many neurons in parallel, but the neurons in such hardware implementations may have very limited precision. To account for the possible limited precision in implementation, { in software} we simulate subthreshold membrane potential decay with 1ms time precision and restrict all neuron parameters { ($V_{thresh}$, $\tfrac{I_l}{C_m}$, and $t_{refr}$ in Table~\ref{Table:NetworkParameters})} to be { unsigned} 8 bit integers {with 1 Least Significant Bit (LSB) corresponding to 1 unit shown in Table~\ref{Table:NetworkParameters}. During simulation, membrane potential is stored as an integer value in units of millivolts}.

The simple behaviour of IF neurons ensures that an output spike can only be elicited by an excitatory input spike, and not by subthreshold membrane potential dynamics in the absence of excitatory input. When an input to a neuron arrives, the neuron's new state (membrane potential) can be entirely determined by the time since it was last updated, and its state after the previous update. We therefore need only update a neuron when it receives an input spike (rather than at a constant time interval). Neurons are organized into a hierarchical structure consisting of layers. When an input spike arrives from a lower layer, the update procedure for the neuron is:

\begin{algorithmic}
\If {$t_i-t_{lastspike} < t_{refr}$}    
    \State{$Vm_{i} \leftarrow Vm_{i-1}$}
\Else                                   
\State{$Vm_i \leftarrow  \left\{
  \begin{array}{l l}
    \max{\{Vm_{i-1} - \frac{I_l}{C_m}(t_i-t_{i-1}), 0\}} & \text{if } Vm_{i-1} \geq 0\\
    \min{\{Vm_{i-1} + \frac{I_l}{C_m}(t_i-t_{i-1}), 0\}} & \text{if } Vm_{i-1} <0
  \end{array} \right.$}\\
\State{$Vm_i \leftarrow  Vm_i + \omega_i$}
\EndIf\\

\If {$Vm_i \geq V_{thresh}$} 
\State{$Vm_i \leftarrow 0$}
\State{$t_{lastspike} \leftarrow t_i$}
\State{\textbf{Do}(Generate Output Spike)}
\EndIf
\end{algorithmic}

where $t_i$ is the time at which the $i^{th}$ input spike arrives, $t_{lastspike}$ is the time at which the current neuron last generated an output spike, $t_{refr}$ is the refractory period of the neuron, $Vm_i$ is the membrane voltage after the $i^{th}$ input spike, $I_l$ is the leakage current, $C_m$ is the membrane capacitance, $\omega_i$ is the input weight of the $i^{th}$ input spike, and $V_{thresh}$ is the threshold voltage for the current neuron.

Output spikes from a neuron feed similarly into the layer above, but can also affect neurons within the same layer through lateral connections. When an input is received from a lateral connection, it forces the receiving neuron to reset and enter a refractory period. In practice we implement this by treating the {reset} neuron into thinking it has recently spiked by using the update:
\begin{algorithmic}
\State{$t_{lastspike} \leftarrow t$}
\end{algorithmic}
where $t$ is the current time.

\subsection{Using Spike Timing to Find the Max}\label{sec:TimingMax}
\label{sec:timing}

Jarrett~\emph{et~al.} \cite{BestArchitectureForRecognition} showed in a comparison of object recognition architectures that the top performing algorithms are those with a hierarchical structure incorporating a non-linearity, although some more recent works show similar performance with a single layer of neurons, but at the expense of increased computational complexity and training difficulty \cite{Nielsen2015}. In the case of the popular HMAX \cite{HMAX2007} model, this non-linearity is a maximum operation in the pooling stages (C1 and C2). Finding this maximum requires comparing the responses of all units within the region to be pooled. This maximum value is then passed through to the next layer, irrespective of how large or small the value is. { In other words, the maximum value is passed to the next layer, regardless of its value (so long as it is the maximum).}

\begin{figure}
  \centering
  \includegraphics[width=0.5\textwidth]{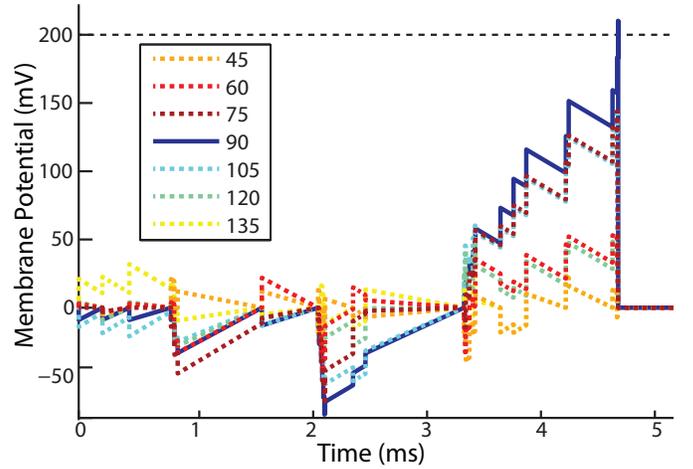}\\
  \caption{Competition between neurons tuned to different orientations when presented with a visual edge oriented at 90 degrees. The neuron tuned to 90 degrees is strongest stimulated causing it to cross spiking threshold first and {reset} all other orientations.}\label{fig:OrientationCompetition}
\end{figure}

In the HFirst architecture we observe which neuron responds first, and judge that neuron to have the maximal response to the stimulus. This is based on two main observations. Firstly, that sharper edges (larger spatial gradients) result in larger temporal contrast \eqref{eq:ImageConstancy}, therefore generating events sooner than less sharp edges. Secondly, the higher the spatial correlation between a neuron's input weights and the spatial pattern of incoming spikes, the stronger it will be activated (see Fig.~\ref{fig:OrientationCompetition}). The strongest activated neuron will cross its spiking threshold before other neurons, thereby providing an indication that its response is strongest. Using this mechanism there is no need to compare neuron responses to each other, rather we simply observe which neuron generated an output first. The first spike from a pooling region can then be used to reset other orientations through lateral {reset connections}, thereby ensuring that non-maximal responses are not propagated through to subsequent layers.

Fig.~\ref{fig:OrientationCompetition} shows how neurons tuned to different orientations will respond when an edge is presented. The neuron tuned to the orientation of the edge (90 degrees, solid line) is strongest activated and crosses the spiking threshold before other neurons (dotted lines). Neurons tuned to orientations similar to the stimulus (75 and 105 degrees) are next strongest activated, but are { reset} by the neuron sensitive to 90 degrees (since it spiked first). Neurons tuned to orientations below 45 degrees and above 135 degrees are not shown to reduce figure clutter.

The ``time to first spike" approach simplifies computation of the max. It indicates which neuron has the strongest response, and through the time at which the spike is elicited it conveys how strong the response is. However, if no neuron was activated strongly enough to generate an output spike, no first-spike is detected and no output spikes are generated. This is an important property ensuring that no computation is performed when there is insufficient activity in the scene. Much like the front end sensor, which represents lack of stimulus (temporal contrast) through a lack of data, HFirst represents the lack of a strong enough neuron activation through a lack of output spikes.

\section{Asynchronous HFirst Architecture}
\label{sec:AsyncArchitecture}

\begin{figure}
  \centering
  \includegraphics[width=0.5\textwidth]{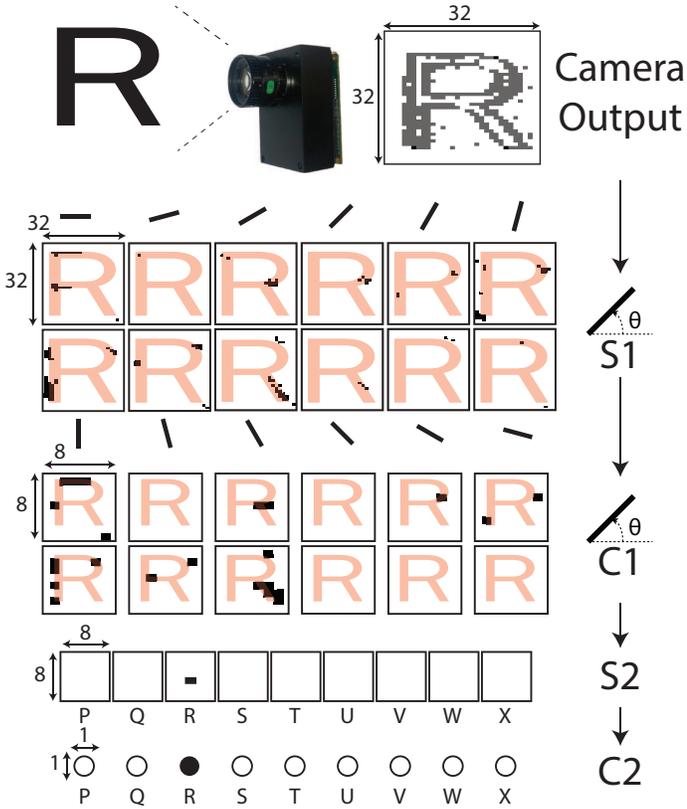}\\
  \caption{The HFirst model architecture, consisting of four layers (S1, C1, S2, C2). Only a 32$\times$32 pixel cropped region of real data extracted from the model is shown to ease visibility while demonstrating recognition of the character `R'. Black dots represent data from the model. The character `R' has been superimposed on top of the S1 and C1 data to aid explanation. The size of the (cropped) data is shown at the left of each layer (Table~\ref{Table:NetworkParameters} shows the sizes for the full model). The S1 layer performs orientation extraction at a fine scale, followed by a pooling operation in C1. Note that due to lateral reset in C1, some S1 responses are blocked (for example, the last three orientations on the bottom row). The S2 layer combines responses from different orientations, but maintains spatial information. The C2 layer pools across all S2 spatial locations, providing only a single output neuron for each character.
  }\label{fig:Architecture}
\end{figure}

HFirst is structured in a similar manner to hierarchical neural models \cite{HMAX2007, masquelier2007unsupervised}, which consist of four layers, named Simple 1 (S1), Complex 1 (C1), Simple 2 (S2), and Complex 2 (C2). In these frame base architectures, cells in simple layers densely cover the scene and respond linearly to their inputs, while cells in complex layers have a non-linear response and only sparsely cover the scene. The layers and manner in which computation is performed in HFirst differs considerably from previous implementation of similar computational models of object recognition in cortex \cite{Riesenhuber:1999, HMAX2007, FeedforwardRapidCategorization}. The Simple layers in HFirst are in fact non-linear due to the use of a spike threshold and binary spike output. In the remainder of this section the form and function of each HFirst layer is described. The same neuron model is used for all layers, but with different parameters and connectivity. The network architecture is shown in Fig.~\ref{fig:Architecture}, and the parameters for each stage are shown in Table~\ref{Table:NetworkParameters}.

\subsection{Layer 1: Gabor Filters}
The S1 layer densely covers the scene with even Gabor filters at 12 orientations. All filters are 7x7 pixels, resulting in 12 filters at each pixel. These filters are designed to pick up sharp edges. Filter kernels are generated with the same equation as in Serre~\emph{et~al.} \cite{HMAX2007}, repeated below for convenience.

\begin{equation}
\label{eq:Original Filters}
\begin{array}{l l}
F_{\theta}(u,v) &= e^{(-\frac{u_0^2+\gamma^2v_0^2}{2\sigma^2})} \cos{(\frac{2\pi}{\lambda}u_0)}\\[8pt]
u_0 &= u\cos{\theta} + v\sin{\theta}\\[8pt]
v_0 &= -u\sin{\theta} + v\cos{\theta}.\\
\end{array}
\end{equation}
where $u$ and $v$ are horizontal and vertical location in pixels. $u_0$ and $v_0$ are used to effect a rotation which orients the filter. Parameters of $\lambda = 5$ and  $\sigma = 2.8$ were used to generate the synaptic weights. $\theta$ varies from 0 to 165 degrees in increments of 15 degrees.

S1 neurons are divided into adjacent non-overlapping 4$\times$4 pixel regions, referred to as S1 units. Each S1 unit feeds into 12 C1 neurons, one for each orientation. C1 neurons have lateral {reset} connections between orientations to perform the max operation discussed in Section~\ref{sec:timing}. C1 neurons use a very low threshold voltage to ensure that a single input spike is sufficient to generate an output spike (provided the neuron is not under refraction).

The refractory period in C1 saves computation by reducing the number of spikes which need to be routed within the architecture. Limiting the firing rate is also important to ensure that no single C1 neuron can fire rapidly enough to single handedly elicit a spike from an S2 neuron.

\begin{table}
  \caption{{Neuron Parameters}}\label{Table:NetworkParameters}
\begin{center}
\begin{tabular}{|c||c|c|c|c|c|}
\hline
Layer               &$V_{thresh}$   &${I_l}/{C_m}$      &$t_{refr}$         & Kernel Size               & Layer Size  \\ \hline
S1                  &200            &50                     &5                  &7$\times$7$\times$1        & 128$\times$128$\times$12     \\ \hline
C1                  &1              &0                      &5                  &4$\times$4$\times$1        & 32$\times$32$\times$12  \\ \hline
S2                  &100-200        &10                     &10                 &8$\times$8$\times$12       & 32$\times$32$\times N_y$    \\ \hline
C2                  &1              &0                      &10                 &32$\times$32$\times 1$     & 1$\times$1$\times N_y$\\ \hline
unit                &mV             &mV/ms                  &ms                 &synapses                   & neurons     \\ \hline
\end{tabular}
\end{center}
\end{table}

\subsection{Layer 2: Template Matching}
\label{sec:template_matching}
S2 neurons densely cover C1 neurons, with each receiving inputs from 8$\times$8 C1 neurons of all orientations. S2 receptive fields are created during a training phase as described below.

A simple activity tracker \cite{goalKeeperRetina} is used to track training objects and compensate for their motion to generate a static 32$\times$32 pixel view of the object. This stabilised view is processed by S1 and C1, and the number of spikes of each orientation originating from each C1 neuron is counted. Note that due to the non-overlapping S1 units, the 32$\times$32 pixel input region feeds into 8$\times$8 C1 neurons, which is the size of an S2 receptive field in HFirst (see Table~\ref{Table:NetworkParameters}).

The counts generated in this manner constitute the synaptic weights (or input kernel) for the S2 neuron sensitive to this object. A separate neuron is required for each object to be recognized. For each neuron, synaptic weights are normalised to have an $l_2$ norm of 100. Finally, since negative spike counts are not possible, all zero valued weights are replaced with inhibitory values (-1) to reduce noise sensitivity. A copy of each trained neuron is then implemented at every location, allowing detection of all trained objects at all locations.

\begin{figure}
  \centering
  \includegraphics[width=0.5\textwidth]{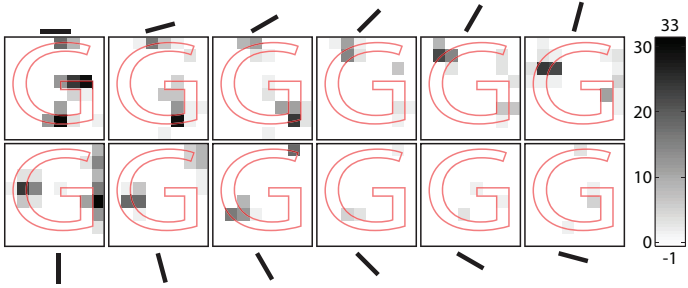}\\
  \caption{The receptive field of an S2 neuron trained to recognize the character `G'. The neuron receives inputs from an 8$\times$8 (x$\times$y) C1 region and from all 12 orientations (orientations are indicated by the oriented blacked bars). Dark regions indicate strong excitatory weights and can be seen to fall along edges of the character wherever edge orientation matches the C1 neuron orientation. Weaker response between 135 and 165 degrees (bottom right) are due to the direction of motion of the character during training (roughly 150 degrees). Motion perpendicular to the direction of motion is required to elicit temporal contrast, as shown in \eqref{eq:ImageConstancy} and discussed in Section~\ref{sec:ATIS}. After normalization the weights in this example range from -1mV to 33mV, indicated by the bar on the right.  }\label{fig:S2Features}
\end{figure}

Fig.~\ref{fig:S2Features} shows an example of a learnt S2 receptive field for recognizing the character `G'. The figure shows how the highest input synapse weights are assigned to locations where the orientation of character's edges match the orientation to which the underlying C1 neurons are tuned.

S2 neuron spikes {reset} all other S2 neurons within an 8x8 region sensitive to other classes of objects, thus implementing the max operation discussed in Section~\ref{sec:TimingMax}. Furthermore, by only {resetting} neurons sensitive to \emph{other} object classes, the detected object class is given a ``head start" in the race to first spike in the nearby region. This can be seen as using the detection to create a prior expectation of detecting that object again nearby.

An optional C2 layer can be used to pool all responses from all S2 locations for classification. The C2 layer is not always used because it discards information regarding the location of the object, which can be particularly useful when multiple objects of interest are simultaneously present in the scene.

\subsection{Classifier}
A basic classifier outputs the soft probabilities for the object belonging to each class. The probability $P(i)$ of an object belonging to class $i$ is calculated as
\begin{equation}
P(i) = \frac{n_{i}}{\sum_i{n_{i}}}
\end{equation}
where $n_i$ is the number of spikes elicited by S2 neurons sensitive to the $i^{th}$ class. When ${\sum_i{n_{i}}}=0$ we assign $P(i) = 0$ for all classes.

If we wish to force the classifier to choose only a single class, we can assign the output class $y$ as
\begin{equation}
y = \max_i({n_{i})}
\end{equation}

We have no neuron to respond to lack of an object in a scene. Lack of an object results in lack of positive detections. This is a fundamental concept of the computing and sensing paradigm we use. Lack of information is not communicated, but is rather represented by a lack of communicated data.

\section{Implementation}
\label{sec:ComputationalRequirements}
In this section we briefly analyse computational requirements. The number of input spikes generated by the front end sensor varies with scene activity and dictates the required computation since neuron updates are only performed when spikes are received. We analyse computation as a function of the number of input and output spikes for each layer. A worst case scenario is used which assumes that a neuron is updated every time it receives a spike (ignoring the refractory period).

A parallelised and pipelined FPGA implementation was programmed to run in real-time on the Opal Kelly XEM6010-LX150 board, which includes a Xilinx Spartan 6 XC6SLX150-2 FPGA. The model operates on a 128$\times$128 pixel input. The implementation runs at a clock frequency of 100MHz and uses internal block RAM without relying on external RAM. The final output of the system consists of S2 output spikes, although access is also provided to spikes from intermediate layers for characterization.

\subsection{S1 and C1: Gabor Filters}
Each input spike in S1 routes to all S1 neurons within a 7$\times$7 pixel region. There are 12 S1 neurons at each pixel location (one per orientation), resulting in 12$\times$7$\times$7 = 588 synapse activations per input spike.

For FPGA implementation, 84 synapses update in parallel, requiring 7 clock cycles to update all 588 synapses, allowing the S1 stage to sustain throughput of 14M events per second.

Each S1 output spike excites a single C1 neuron, and {resets} the 11 C1 neurons sensitive to other orientations, resulting in 12 C1 synapse activations per S1 output spike. C1 updates all 12 synapses in parallel and can process 25M input events per second.

\subsection{S2 and C2: Template Matching}
Each input spike to S2 routes to all S2 neurons within an 8$\times$8 region. If $N_y$ denotes the number of classes to be classified, then there will be $N_y$ neurons at each location, and each input spike will activate $N_y\times$8$\times$8 = 64$N_y$ input synapses.
Each S2 output spike {resets} all S2 neurons lying within an 8$\times$8 region around where the spike originated. So, for every S2 output spike $N_y\times$8$\times$8 = 64$N_y$ S2 lateral {reset} synapses are activated.

The FPGA implementation of S2 can update $N_y$ neurons in parallel, requiring 64 clock cycles to process each input or output spike. The number of C2 input synapses activated is equal to the number of S2 neuron output spikes. The C2 stage is optional and not implemented in FPGA.

{In HFirst there are no zero valued synapses in S2. Synapses which are not activated during training are assigned an inhibitory synaptic weight of -1 (see Section~\ref{sec:template_matching}). The number of synapses in S2 could be significantly reduced by instead assigning a weight of 0 to these synapses and optimizing them out of the model. However, such an optimization would introduce significant additional complexity in pipelining for the FPGA implementation. The FPGA implementation benefits far more from the simplified pipelining which results from having a dense regular connection structure where all synapses are implemented. This connection structure is also more general, allowing the synaptic weights to be easily reprogrammed.

The regular connection structure also saves memory by ensuring that when a neuron is updated, all co-located neurons will also be updated. Updating all co-located neurons simultaneously allows us to store only a single time value to indicate when all neurons at that location were updated, rather than storing a separate time value to indicate when each individual neuron was last updated (Section~\ref{sec:neuron_model} shows how the time value is used in the neuron update). This memory saving is important because memory availability is the limiting factor in scaling the model to higher resolution, as shown in the next section.
}

\begin{table}
  \caption{Required Computation and Resources}\label{Table:Hardware Resources}
\begin{center}
\begin{tabular}{|l||r|r|r|r|}
\hline
\textbf{Stage}                          &\textbf{S1}    &\textbf{C1}    &\textbf{S2}    &\textbf{C2}\\ \hline
\textbf{Synapse updates per event}      &588            &12             &64$N_y$        &1          \\ \hline
\textbf{Throughput events/sec}          &14M            &25M            &1M             &100M       \\ \hline
\textbf{DSP blocks}                     &16             &0              &1              &0          \\ \hline
\textbf{Block RAM}                      &128            &2              &$N_y$+1        &0          \\ \hline
\end{tabular}
\end{center}
\end{table}

\subsection{Scaling to higher resolution}
\label{sec:Scaling}
When scaling to higher resolutions two main factors need to be considered: memory requirements, and computational requirements. Required memory scales linearly with the number of neurons in the model, which in turn scales linearly with the number of input pixels. Computational requirements scale linearly with the input event rate.

With 36 classes ($N_y~=~36$), 167 Block RAMs are used for HFirst (see~Table~\ref{Table:Hardware Resources}), plus an additional 10 for pipeline FIFOs and USB IO, resulting in a total of 177 of the available 268 Block RAMs being used for 128$\times$128 pixel input resolution.

{Digital Signal Processing (DSP) blocks are blocks within the FPGA containing dedicated hardware for performing multiplication and addition. The number of multiplications which can be performed per second is a limiting factor in many algorithms, particularly for visual processing algorithms which compute kernel responses using convolution. Optimization of these algorithms typically involves optimizing memory access and pipelining to maximise utilization of hardware multipliers (see \cite{Orchard2013} for an example). High end GPUs and FPGAs contain thousands of hardware multipliers.

In HFirst only 17 of our FPGA's 180 DSP blocks are used and these 17 DSP blocks are only utilized a small percentage of the time due to the temporal sparsity of the AER data. For HFirst, internal FPGA memory is the limiting resource when increasing resolution. Internal memory requirements scale with the input sensor resolution, while the number of DSP blocks required will scale with maximum sustained input event rate the model is required to handle.

The current FPGA implementation can handle a sustained 14Meps (events per second) input event rate, while bursts of up to 100Meps  (limited by FPGA clock speed of 100MHz) can be handled for durations up to 5$\mu$s (limited by FIFO buffer depth). Larger FIFO buffers can be used, but are unnecessary. At 128$\times$128 resolution, event rates for typical scenes are around 1Meps. The latest ATIS can generate events at a peak rate of 25Meps, and sustain a maximum rate of 15Meps at 304$\times$240 pixel resolution. 14Meps is therefore a very high rate for 128$\times$128 pixel resolution. Using additional DSP blocks, the maximum sustainable event rate can be increased by 1Meps per block used.
}

\subsection{Power Consumption}

The FPGA board on which HFirst was implemented also performs other tasks in parallel as part of normal operation of the ATIS sensor (powering and controlling the ATIS, as well as interfacing to a host PC). Implementing HFirst in addition to the other tasks on the FPGA increases power consumption by 150mW for static scenes (little to no processing happening), and by a further 100mW for the the highest activity scene we could generate. We therefore estimate HFirst power consumption to be between 150mW and 250mW depending on scene activity. These measurements are done at the board's power supply and include losses due to inefficiencies in the onboard switching regulators.

\section{Testing}
\label{sec:Testing}
\begin{figure}
  \centering
  \includegraphics[width=0.5\textwidth]{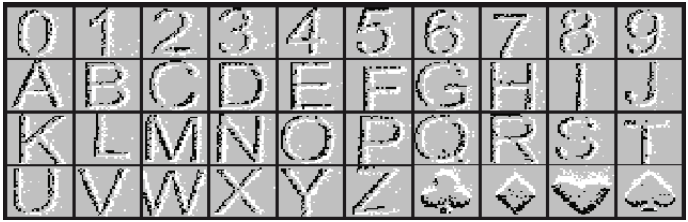}\\
  \caption{Examples of the stabilised characters and cards pip views used for training. Each example measures 32$\times$32 pixels and shows 1.7ms of data.}\label{fig:Examples}
\end{figure}

HFirst was tested on two tasks. The first consists of recognizing pips on poker cards as they are shuffled in front of the sensor. The poker card task has been previously tackled \cite{BernabeConvNet} and was chosen to provide a direct comparison with previously published works. The second task is a simulated reading task in which characters are recognized as they move across the field of view using the test setup shown in Fig.~\ref{fig:TestSetup}. Examples of recordings used for each task are shown in Fig.~\ref{fig:Examples}. For both tasks, HFirst was implemented in Matlab simulation, coupled with a reconfigurable C++ function for increased speed.

\subsection{Poker cards}
For the poker card task data was provided by Linares-Barranco \cite{BernabeConvNet} who captured the data using the sensitive DVS sensor \cite{BernabeRetina}. The dataset consists of 10 examples for each of the 4 card types (spades, hearts, diamonds, and clubs). For each of 10 different trials, non-overlapping test and training sets were chosen such that each contained 5 examples of each pip. For each pip in the training set, all 5 examples were concatenated into a single sequence from which the S2 layer kernel was generated. To provide a close comparison with the previously published task, we also tested on the stabilised and extracted pips.

Additional tests were performed in which {lateral reset connections} were removed from the model to investigate the value of the timing approach to computing the max. Finally, the advantage of having orientation extraction and pooling in S1 and C1 were investigated by bypassing these stages.

\subsection{Character Recognition}
\begin{figure}
  \centering
  \includegraphics[width=0.5\textwidth]{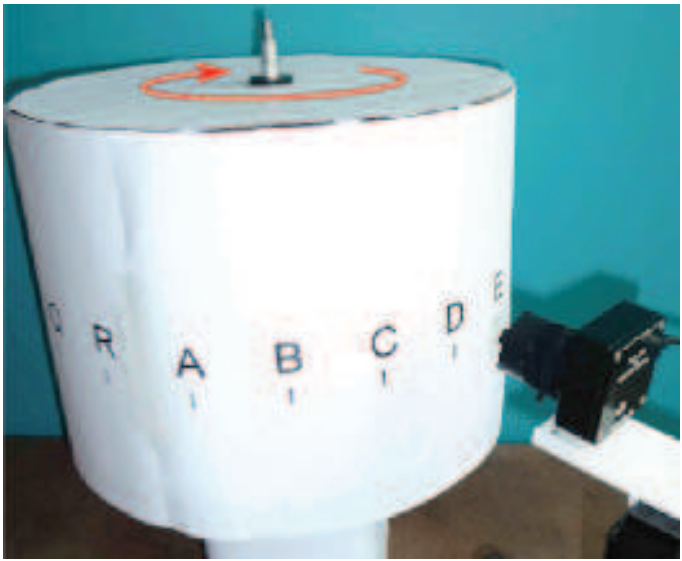}\\
  \caption{The test setup used to acquire the character dataset, consisting of a motorised rotating barrel covered with printed letters viewed by a DVS \cite{DVStobi}.}\label{fig:TestSetup}
\end{figure}

36 characters (0-9 and A-Z) were printed on the surface of a barrel which was rotated at 40rpm while viewed by the DVS \cite{DVStobi} as shown in Fig.~\ref{fig:TestSetup}. Data was recorded over two full rotations of the barrel, thereby providing two recordings for each character. For each of 10 trials, non-overlapping test and training sets were randomly chosen such that every character appears once in each set. Training and testing was then performed using an automated script.

Training of the second layer of HFirst is performed on a stabilised view of a moving object, and therefore requires knowledge of the object location, which is acquired through tracking. However, for testing we use moving sequences instead of stabilised views, removing the need for tracking.

As with the card task, The character recognition task was also used to investigate the advantages of using {reset connections} for max computation, and of performing orientation extraction and pooling in S1 and C1 respectively.

{ Further testing was performed on the characters to show that HFirst can detect multiple objects simultaneously present in the scene, and to investigate the impact of timing jitter introduced during training and testing.

Finally the importance of precise timing was investigated by artificially altering spike times in the recordings and observing the effect on HFirst accuracy.}

\section{Results}
\label{sec:Results}
Results from testing are summarised in Table~\ref{Table:Results Summary}, and discussed in the sections below. The S1 and C1 columns show the total number of activated synapses in each of these layers. For S2, the S2 and S$2_{rst}$ columns show the number of activated feedforward (from C1) and {lateral reset} synapses respectively.

\begin{table}
  \caption{Detection Accuracy and Required Computation}\label{Table:Results Summary}
\begin{center}
\begin{tabular}{|l||l|r|r|r|r|}
\hline
\multirow{2}{*}{\textbf{Task}} &\textbf{Accuracy }     &\multicolumn{4}{c|}{\textbf{Input Synapse Activations}} \\
                               &\textbf{\%}            &\textbf{S1}    & \textbf{C1}   &\textbf{S2}  &\textbf{S$2_{rst}$}\\ \hline
HFirst Cards                                &                       &               &              &              &\\
~~Full model                                &$97.5\pm3.5 $          &2.6M           &10k           &19k           &710\\
~~No S1, C1 {reset}              &$51.6\pm4.4$           &2.6M           &3.8k          &127k          &79k\\
~~No S2, C2 {reset}              &$72.3\pm3.8$           &2.6M           &10k           &19k           &-\\
~~No {reset}                     &$24.9\pm0.1 $          &2.6M           &3.8k          &127k          &-\\
~~Bypass S1                                 &$49.1\pm5.4$           &-              &4.3k          &37k           &20k\\
~~Bypass S1 and C1                          &$60.7\pm3.5$           &-              &-             &1.1M          &333k\\
CNN Cards                                   &                       &               &              &              &   \\
~~Spiking \cite{BernabeConvNet}             &91.6                   &-              &-             & -            & -\\
~~Frame based \cite{BernabeConvNet}         &95.2                   &-              &-             & -            & -\\ \hline \hline
HFirst Characters                           &                       &               &              &              &\\
~~Full model                                &$84.9\pm1.9$           &8.4M           &40k           &720k          &159k\\
~~No S1, C1 {reset}              &$70.4\pm5.8$           &8.4M           &8.3k          &2.4M          &309k\\
~~No S2, C2 {reset}              &$56.7\pm0.9$           &8.4M           &40k           &720k          &-\\
~~No {reset}                     &~$4.6\pm0.1$           &8.4M           &8.3k          &2.4M          &-\\
~~Bypass S1                                 &$31.2\pm4.1$           &-              &14k           &1.6M          &1.1M\\
~~Bypass S1 and C1                          &$81.4\pm3.8$           &-              &-             &33M           &32M\\  \hline
\end{tabular}
\end{center}
\end{table}

\subsection{Cards}
HFirst classified the stabilised and extracted card pips with an accuracy of 97.5\%$\pm$3.5\% using an S2 threshold of 150mV. Chance for this task is 25\%. The average duration of a test example was 23ms, and consisted of 4.3k input spikes, which elicited 73 C1, and 2.8 S2 spikes. The S1/C1 and S2/C2 layers took on average 102ms and 0.7ms respectively per example to simulate in Matlab using a single thread on an Intel Xeon X5675 processor running at 3.07GHz. The FPGA implementation simulates the network in real-time, with latency $\leq2\mu s$ in response to incoming events.

Removing lateral {reset} in the first layer decreases recognition accuracy to 51.6\%$\pm$4.4\%, while removing lateral {reset} connections in the second layer decreases recognition accuracy to 72.3\%$\pm$3.8\%, and removing lateral {reset connections} in both layers reduces recognition accuracy to chance levels, while increasing the average number of spikes elicited to 309 and 66 for C1 and S2 respectively. These results suggest that using the first spike mechanism improves performance, both in terms of computational efficiency, and in terms of recognition accuracy.

For the card classification task which only has four output classes, bypassing the first layers reduces the required computation at the cost of recognition accuracy.

\subsection{Characters}
HFirst classified the moving letters with an accuracy of 84.9\%$\pm$1.9\% using an S2 threshold of 200mV. Chance for this task is 2.8\%. The average duration of a test example was 112ms, and consisted of 14k input spikes, which elicited 313 C1, and 27 S2 spikes on average. The S1/C1 and S2/C2 layers took on average 365ms and 28ms respectively per example to simulate in Matlab using a single thread on an Intel Xeon X5675 processor running at 3.07GHz. As with the card pip task, the FPGA implementation easily runs in real-time with latency $\leq2\mu s$.

Next we investigated the effects of bypassing the first layers of HFirst and performing template matching directly on the input events. This modification resulted in an accuracy of 81.4\%$\pm$3.8\%, which is not too different from the performance of the full model. However, bypassing the S1 and C1 layers also increases the required computation significantly, suggesting that performing orientation extraction and pooling in S1 and C1 is actually more computationally efficient. The same is not true for the cards task where only 4 classes are present, but is true whenever 10 or more output classes are required. This increased computational requirement is also obvious when observing the time taken for simulation, which increased by 50 fold to an average of 19.7 seconds per example in Matlab.

\subsection{Detecting Multiple Objects Simultaneously}
\begin{figure}
  \centering
  \includegraphics[width=0.5\textwidth]{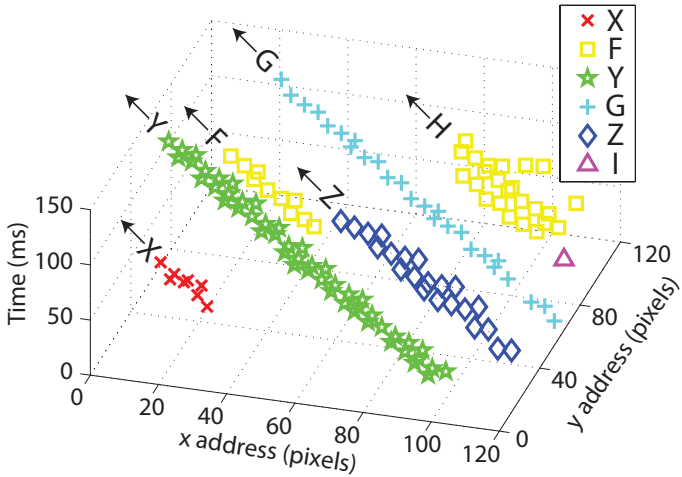}\\
  \caption{HFirst S2 layer spikes (indicated by markers) over a 150ms time period in response to the character data. This figure shows the ability of HFirst to detect multiple characters in the scene simultaneously. Both location of the objects and their class are indicated by S2 spikes. The `X', `F', `Y', and `G' characters are correctly detected, but the character  `H' is misclassified, being mistaken for an `I' or `F' at different times.}\label{fig:AllCharactersOutput}
\end{figure}

\begin{figure}
  \centering
  \includegraphics[width=0.5\textwidth]{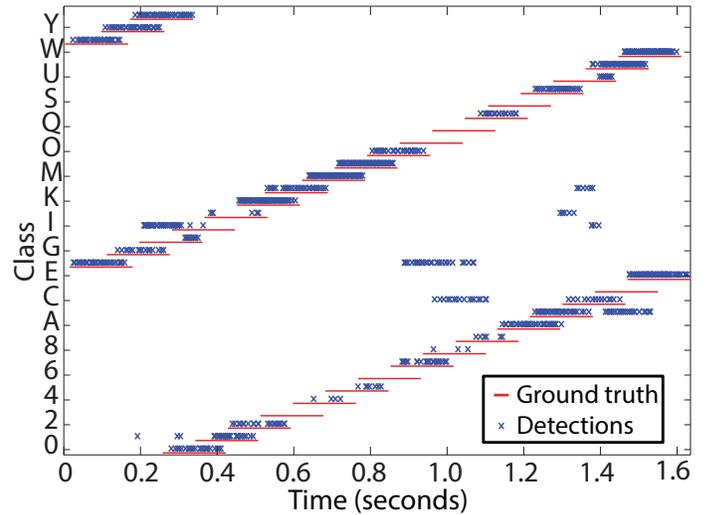}\\
  \caption{Detection of characters for a single rotation of the barrel. Only every second character is labelled on the vertical axis to reduce clutter. Red lines indicate when each character is present in the visual field, while blue crosses mark detections made by HFirst. Note that up to 4 characters are present in the scene at any one time.}\label{fig:AllSequence}
\end{figure}

{
After testing the model performance on individual characters, we verified that it can detect multiple characters simultaneously present in the scene. Fig.~\ref{fig:AllCharactersOutput} shows 150ms worth of S2 outputs with multiple characters simultaneously visible in the scene. The S2 responses indicate both the object class and location. In this example the letters `X', `F', `Y', and `G' are all accurately detected as they pass across the scene. Later, the letters `Z' and `H' enter the scene. The `Z' is accurately detected, but the `H' is erroneously detected as an `F' and `I' at different points in time. The 1 in 6 error for these characters is in agreement with the 84.9\%$\pm$1.9\% accuracy reported overall.

Fig.~\ref{fig:AllSequence} shows output detections for a single full rotation of the barrel, comparing the times at which letters were detected (or missed) to the ground truth of when they were present in the scene.
}

\subsection{Effect of Timing Jitter}
{
In the front end AER sensor, the latency of pixel responses and of the AER readout can vary, resulting in timing jitter in the spikes feeding into S1. All of our tests are performed on real recordings and therefore include some jitter. In order to investigate the effect of increased timing jitter on the model, we artificially added additional jitter to the recordings used for training and testing.
{
{
Jitter times for each spike were randomly chosen from a Gaussian distribution and the effect of varying the standard deviation of the distribution is shown in Fig.~\ref{fig:Jitter}. Changing the mean of the Gaussian distribution adds a constant time offset to all spikes and has no effect on accuracy. The accuracy for each standard deviation value is again obtained as the mean of 10 random test and training splits performed on the character database. Two tests were run, in the first test additional jitter was introduced in the training data (Fig.~\ref{fig:Jitter}a) and the test data was left unaltered. In the second test (Fig.~\ref{fig:Jitter}b) the training data was left unaltered and additional jitter was introduced only in the test data.

Training is performed on tracked and stabilized views of the characters, thus for the purposes of training, the characters appears static. HFirst can therefore tolerate high timing jitter because even when a spike's time is changed, it will still occur in the correct location relative to the center of the character. Accuracy drops off significantly only when the standard deviation of the jitter exceeds 100ms, which is comparable to the length of the recording itself (112ms).

Recognition is performed on moving views of the characters which are crossing the field of view at roughly 1 pixel/ms. Delaying a spike by even a few milliseconds (Fig.~\ref{fig:Jitter}b) will cause the spike to occur in the wrong location relative to the center of the character (because the character center will have moved during the delay period). Therefore, even a few milliseconds of timing jitter will cause a significant decrease in recognition accuracy.

\begin{figure}
  \centering
  \includegraphics[width=0.5\textwidth]{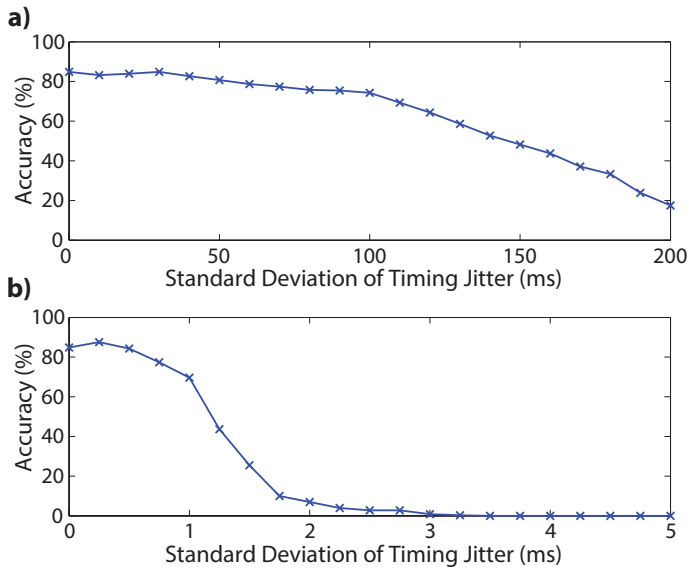}\\
  \caption{The effect of timing noise on recognition accuracy for the character recognition task. Adding Gaussian noise to the stabilized training data (a) has little effect on accuracy because even when delayed, spikes occur in the correct location relative to the character center. Accuracy drops off significantly only when the timing jitter is large enough to cause the training data spikes to be too spread in time. Adding even a small degree of Gaussian noise to the moving characters used for testing (b) causes accuracy to drop off significantly because by the time the delayed (jittered) spikes arrive at the S1 inputs, the character has already moved on to a new location.}\label{fig:Jitter}
\end{figure}
}

\section{Discussion}
\label{sec:Discussion}
In this paper we have described a spiking neural network for visual recognition dubbed ``HFirst". 
HFirst exploits timing information in the incoming visual events to implement a time-to-first spike operation as a temporal Winner-Take-All (WTA) operation with lateral {reset} to block responses from other neurons in the same pooling area. Computationally, this temporal WTA is significantly simpler than the MAX operation typically used in hierarchical models.

HFirst operates on change detection data from AER sensors. Each pixel in these sensors adapts individually to ambient lighting conditions, which to a large extent removes dependence on lighting conditions. This removes the need for normalization of oriented Gabor responses in HFirst, which is another computationally intensive task (division) required by the standard HMAX model and other CNN implementations.

Thus far HFirst has been tested on simple objects, and neurons in the second layer of HFirst directly detect the presence of these objects, allowing HFirst to simultaneously detect multiple objects in the scene, which is not typically possible with CNNs.

Masquelier~\emph{et~al.} \cite{masquelier2007unsupervised} used STDP to learn more complex features, and a powerful Radial Basis Function (RBF) classifier which allows recognition of more complex objects (motorcycles and faces from Caltech 101). Their approach used STDP to extract features with high correlation between training examples, even though these features appear at different locations. This removes the need to precisely track and stabilise a view of an object for training. However, the model only operates on static images, removing the problem of moving stimuli, and objects are already centered in the Caltech 101 database (although features do not always appear at the same location). A second major difference is that HFirst operates continuously, whereas Masquelier~\emph{et~al.} present images to their model sequentially, requiring the system to be reset before each image presentation.

In a recent PAMI paper, Perez-Carrasco~\emph{et~al.} \cite{BernabeConvNet} reported an accuracy ranging from 90.1\% to 91.6\% for the card pip task using a five layer spiking CNN. They kindly provided us with their data and for the same task we report accuracy of 97.5\%$\pm$3.5\%. However, we compute accuracy differently to Perez-Carrasco~\emph{et~al.}. Their CNN implementation includes separate ``positive" and ``negative" responses to represent the presence or absence for each object, and both these responses are used in their calculation of accuracy. HFirst has no ``negative" responses, which prevents us from using the same equation. Instead, HFirst provides only positive responses, and does not respond when no objects of interest are present in the scene. Nevertheless, if we consider a lack of response from a neuron to be a ``negative" response, then we can use the same equation. Doing so marginally increases our accuracy to 98.8\%$\pm$1.9\% because correct ``negative" responses are rewarded, even when ``positive" responses are incorrect.

The card pip task was also used to investigate the benefits of including lateral {reset}, by showing that removal of lateral {reset connections} in the first, second, or both layers consistently reduces recognition accuracy, while simultaneously increasing computational requirements.

Given the high accuracy of the full HFirst model on the card pip recognition task, a second more difficult character recognition task was constructed and was also used to investigate the benefits of a multi-layer model. Bypassing the first layer decreased accuracy from 84.9\%$\pm$1.9\% to 81.4\%$\pm$3.8\%, suggesting that the first layer increases recognition accuracy. Perhaps more importantly, the first layer significantly reduces computational requirements for the character recognition task. The same was not true for the card recognition task because it consists of very few classes (4), but as the number of classes increases, so does the number of neurons in S2, therefore making it more important to have the S1 and C1 layer to reduce the number of spikes reaching S2.

The leaky integrate and fire neurons used in HFirst essentially perform coincidence detection on input spikes arriving in a specific spatial pattern. A neuron will only generate an output spike if enough input spikes matching this pattern are received within a sufficiently short time period. Under ideal circumstances (no noise), the projection of an object moving between two points on the focal plane will generate the same number of spikes from the AER sensor, regardless of the speed of the object. However, the speed of the object will determine the time period over which these spikes are generated, with slow moving objects not generating spikes at a high enough rate to elicit a response from HFirst layer~1 neurons, but this can be overcome through active sensing, by using a small motion or vibration of the sensor to elicit an egomotion induced velocity on the image plane.

\section{Conclusion}
We have presented an HMAX inspired hierarchical SNN architecture for visual object recognition dubbed `HFirst'. The architecture uses an SNN to exploit the precise spike timing provided by asynchronous change detection vision sensors to simplify implementation of a non-linear pooling operation commonly used in bio-inspired recognition models. HFirst obtains the best reported accuracy on a card pip recognition test and results for a second, far more difficult character recognition task have also been presented. The low computational requirements of the HFirst model allow for real time implementation on an Opal Kelly XEM6010 FPGA board which interfaces directly with the vision sensor, and is both narrower and shorter than a credit card in size.

\section*{Acknowledgment}
The authors thank Bernabe Linares-Barranco for supplying the card pip data, discussions at the Telluride Neuromorphic Cognition Engineering Workshop for helping to formulate these ideas, and the Merlion programme of the Institut Francais de Singapour for facilitating ongoing collaboration on this project.

\bibliographystyle{IEEEtran}

\bibliography{TPAMI-2013-11-0831}

\begin{thebibliography}{10}
\providecommand{\url}[1]{#1}
\csname url@samestyle\endcsname
\providecommand{\newblock}{\relax}
\providecommand{\bibinfo}[2]{#2}
\providecommand{\BIBentrySTDinterwordspacing}{\spaceskip=0pt\relax}
\providecommand{\BIBentryALTinterwordstretchfactor}{4}
\providecommand{\BIBentryALTinterwordspacing}{\spaceskip=\fontdimen2\font plus
\BIBentryALTinterwordstretchfactor\fontdimen3\font minus
  \fontdimen4\font\relax}
\providecommand{\BIBforeignlanguage}[2]{{%
\expandafter\ifx\csname l@#1\endcsname\relax
\typeout{** WARNING: IEEEtran.bst: No hyphenation pattern has been}%
\typeout{** loaded for the language `#1'. Using the pattern for}%
\typeout{** the default language instead.}%
\else
\language=\csname l@#1\endcsname
\fi
#2}}
\providecommand{\BIBdecl}{\relax}
\BIBdecl

\bibitem{SpikeTimingGawne01081996}
T.~J. Gawne, T.~W. Kjaer, and B.~J. Richmond, ``Latency: another potential code
  for feature binding in striate cortex,'' \emph{Journal of Neurophysiology},
  vol.~76, no.~2, pp. 1356--1360, 1996.

\bibitem{SpikeTimingGreschner01122006}
M.~Greschner, A.~Thiel, J.~Kretzberg, and J.~Ammermüller, ``Complex
  spike-event pattern of transient on-off retinal ganglion cells,''
  \emph{Journal of Neurophysiology}, vol.~96, no.~6, pp. 2845--2856, 2006.

\bibitem{masters1993practical}
T.~Masters, \emph{Practical neural network recipes in C++}.\hskip 1em plus
  0.5em minus 0.4em\relax Morgan Kaufmann, 1993.

\bibitem{egmont2002image}
M.~Egmont-Petersen, D.~de~Ridder, and H.~Handels, ``Image processing with
  neural networks: a review,'' \emph{Pattern recognition}, vol.~35, no.~10, pp.
  2279--2301, 2002.

\bibitem{bishop1995neural}
C.~M. Bishop, \emph{Neural networks for pattern recognition}.\hskip 1em plus
  0.5em minus 0.4em\relax Oxford university press, 1995.

\bibitem{NeuFlowISCAS2010}
C.~Farabet, B.~Martini, P.~Akselrod, S.~Talay, Y.~LeCun, and E.~Culurciello,
  ``Hardware accelerated convolutional neural networks for synthetic vision
  systems,'' \emph{IEEE Int. Symp. Circuits and Systems}, pp. 257--260, Jun
  2010.

\bibitem{NeuroGrid}
B.~Benjamin, P.~Gao, E.~McQuinn, S.~Choudhary, A.~R. Chandrasekaran, J.-M.
  Bussat, R.~Alvarez-Icaza, J.~V. Arthur, P.~A. Merolla, and K.~Boahen,
  ``Neurogrid: A mixed-analog-digital multichip system for large-scale neural
  simulations,'' \emph{Proc. IEEE}, vol. 102, no.~5, pp. 699--716, May 2014.

\bibitem{Merolla2014}
P.~A. Merolla, J.~V. Arthur, R.~Alvarez-Icaza, A.~S. Cassidy, J.~Sawada,
  F.~Akopyan, B.~L. Jackson, N.~Imam, C.~Guo, Y.~Nakamura, B.~Brezzo, I.~Vo,
  S.~K. Esser, R.~Appuswamy, B.~Taba, A.~Amir, M.~D. Flickner, W.~P. Risk,
  R.~Manohar, and D.~S. Modha, ``A million spiking-neuron integrated circuit
  with a scalable communication network and interface,'' \emph{Science}, vol.
  345, no. 6197, pp. 668--673, Aug. 2014.

\bibitem{IFATGoldberg2001781}
D.~H. Goldberg, G.~Cauwenberghs, and A.~G. Andreou, ``Probabilistic synaptic
  weighting in a reconfigurable network of {VLSI} integrate-and-fire neurons,''
  \emph{Neural Networks}, vol.~14, no. 6–7, pp. 781 -- 793, 2001.

\bibitem{HiAER}
J.~Park, T.~Yu, C.~Maier, S.~Joshi, and G.~Cauwenberghs, ``Live demonstration:
  Hierarchical address-event routing architecture for reconfigurable large
  scale neuromorphic systems,'' in \emph{IEEE Int. Symp. Circuits and Systems},
  2012, pp. 707--711.

\bibitem{BrainScales}
J.~Schemmel, A.~Grubl, S.~Hartmann, A.~Kononov, C.~Mayr, K.~Meier, S.~Millner,
  J.~Partzsch, S.~Schiefer, S.~Scholze, R.~Schuffny, and M.~Schwartz, ``Live
  demonstration: A scaled-down version of the brainscales wafer-scale
  neuromorphic system,'' in \emph{IEEE Int. Symp. Circuits and Systems}, 2012,
  pp. 702--702.

\bibitem{Spinnaker}
E.~Painkras, L.~Plana, J.~Garside, S.~Temple, F.~Galluppi, C.~Patterson,
  D.~Lester, A.~Brown, and S.~Furber, ``{SpiNNaker}: {A} 1-{W} 18-core
  system-on-chip for massively-parallel neural network simulation,''
  \emph{Solid-State Circuits, IEEE Journal of}, preprint, 2013.

\bibitem{NENGO}
C.~Eliasmith and C.~H. Anderson, \emph{Neural engineering: Computation,
  representation, and dynamics in neurobiological systems}.\hskip 1em plus
  0.5em minus 0.4em\relax MIT Press, 2004.

\bibitem{Brian}
D.~F.~M. Goodman and R.~Brette, ``Brian: a simulator for spiking neural
  networks in python,'' \emph{Front. Neuroinform.}, vol.~2, no.~5, 2008.

\bibitem{PYNN}
A.~P. Davison, D.~Br{\"u}derle, J.~Eppler, J.~Kremkow, E.~Muller, D.~Pecevski,
  L.~Perrinet, and P.~Yger, ``Pynn: a common interface for neuronal network
  simulators,'' \emph{Front. Neuroinform.}, vol.~2, 2008.

\bibitem{HumanBrainProject}
H.~Markram, K.~Meier, T.~Lippert, S.~Grillner, R.~Frackowiak, S.~Dehaene,
  A.~Knoll, H.~Sompolinsky, K.~Verstreken, J.~DeFelipe \emph{et~al.},
  ``Introducing the human brain project,'' \emph{Procedia Computer Science},
  vol.~7, pp. 39--42, 2011.

\bibitem{BRAINusa}
A.~P. Alivisatos, M.~Chun, G.~M. Church, R.~J. Greenspan, M.~L. Roukes, and
  R.~Yuste, ``The brain activity map project and the challenge of functional
  connectomics,'' \emph{Neuron}, vol.~74, no.~6, pp. 970--974, 2012.

\bibitem{OctopusRetina}
E.~Culurciello, R.~Etienne-Cummings, and K.~Boahen, ``A biomorphic digital
  image sensor,'' \emph{Solid-State Circuits, IEEE Journal of}, vol.~38, no.~2,
  pp. 281--294, 2003.

\bibitem{OriginalRetina}
K.~A. Zaghloul and K.~Boahen, ``Optic nerve signals in a neuromorphic chip i:
  Outer and inner retina models,'' \emph{Biomedical Engineering, IEEE
  Transactions on}, vol.~51, no.~4, pp. 657--666, 2004.

\bibitem{mahowald1994analog}
M.~Mahowald, \emph{An analog VLSI system for stereoscopic vision}, ser. Kluwer
  international series in engineering and computer science.\hskip 1em plus
  0.5em minus 0.4em\relax Kluwer Academic Publishers, 1994.

\bibitem{DVStobi}
P.~Lichtsteiner, C.~Posch, and T.~Delbruck, ``A 128x128 120 d{B} 15 us latency
  asynchronous temporal contrast vision sensor,'' \emph{IEEE J. Solid-State
  Circuits}, vol.~43, no.~2, pp. 566--576, Feb 2008.

\bibitem{BernabeRetina}
T.~Serrano-Gotarredona and B.~Linares-Barranco, ``A 128x128 1.5\% contrast
  sensitivity 0.9\% fpn 3 us latency 4 mw asynchronous frame-free dynamic
  vision sensor using transimpedance preamplifiers,'' \emph{Solid-State
  Circuits, IEEE Journal of}, vol.~48, no.~3, pp. 827--838, 2013.

\bibitem{ATIS}
C.~Posch, D.~Matolin, and R.~Wohlgenannt, ``A qvga 143 db dynamic range
  frame-free pwm image sensor with lossless pixel-level video compression and
  time-domain cds,'' \emph{IEEE J. Solid-State Circuits}, vol.~46, no.~1, pp.
  259--275, Jan 2011.

\bibitem{ReviewDelbruck}
T.~Delbruck, B.~Linares-Barranco, E.~Culurciello, and C.~Posch,
  ``Activity-driven, event-based vision sensors,'' \emph{IEEE Int. Symp.
  Circuits and Systems}, pp. 2426--2429, Jun 2010.

\bibitem{folowosele2008real}
F.~Folowosele, R.~J. Vogelstein, and R.~Etienne-Cummings, ``Real-time silicon
  implementation of v1 in hierarchical visual information processing,'' in
  \emph{Biomedical Circuits and Systems Conference, 2008. BioCAS 2008.
  IEEE}.\hskip 1em plus 0.5em minus 0.4em\relax IEEE, 2008, pp. 181--184.

\bibitem{Vogelstein05saliency}
R.~J. Vogelstein, U.~Mallik, E.~Culurciello, G.~Cauwenberghs, and
  R.~Etienne-cummings, ``Saliency-driven image acuity modulation on a
  reconfigurable silicon array of spiking neurons,'' in \emph{Advances in
  Neural Information Processing Systems}.\hskip 1em plus 0.5em minus
  0.4em\relax MIT Press, 2005, pp. 1457--1464.

\bibitem{vogelstein2007multichip}
R.~J. Vogelstein, U.~Mallik, E.~Culurciello, G.~Cauwenberghs, and
  R.~Etienne-Cummings, ``A multichip neuromorphic system for spike-based visual
  information processing,'' \emph{Neural computation}, vol.~19, no.~9, pp.
  2281--2300, 2007.

\bibitem{IFATjacob}
R.~J. Vogelstein, U.~Mallik, J.~T. Vogelstein, and G.~Cauwenberghs,
  ``Dynamically reconfigurable silicon array of spiking neurons with
  conductance-based synapses,'' \emph{Neural Networks, IEEE Transactions on},
  vol.~18, no.~1, pp. 253--265, 2007.

\bibitem{3DAER1}
S.~Schraml and A.~Belbachir, ``A spatio-temporal clustering method using
  real-time motion analysis on event-based 3d vision,'' in \emph{Computer
  Vision and Pattern Recognition Workshops (CVPRW), 2010 IEEE Computer Society
  Conference on}.\hskip 1em plus 0.5em minus 0.4em\relax IEEE, 2010, pp.
  57--63.

\bibitem{3DAER2}
P.~Rogister, R.~Benosman, S.-H. Ieng, P.~Lichtsteiner, and T.~Delbruck,
  ``Asynchronous event-based binocular stereo matching,'' \emph{IEEE Trans.
  Neural Netw.}, vol.~23, no.~2, pp. 347--353, 2012.

\bibitem{3DAER3}
R.~Benosman, P.~Rogister, C.~Posch \emph{et~al.}, ``Asynchronous event-based
  hebbian epipolar geometry,'' \emph{IEEE Trans. Neural Netw.}, vol.~22,
  no.~11, pp. 1723--1734, 2011.

\bibitem{benosman2012asynchronousFLOW}
R.~Benosman, S.-H. Ieng, C.~Clercq, C.~Bartolozzi, and M.~Srinivasan,
  ``Asynchronous frameless event-based optical flow,'' \emph{Neural Networks},
  vol.~27, pp. 32--37, 2012.

\bibitem{Orchard2014b}
G.~Orchard and R.~Etienne-cummings, ``{Bio-inspired Visual Motion
  Estimation},'' \emph{Proc. IEEE}, vol. 102, no.~10, pp. 1520 -- 1536, 2014.

\bibitem{goalKeeperRetina}
T.~Delbruck and P.~Lichtsteiner, ``Fast sensory motor control based on
  event-based hybrid neuromorphic-procedural system,'' \emph{IEEE Int. Symp.
  Circuits and Systems}, pp. 845--848, May 2007.

\bibitem{BernabeConvNet}
J.~Perez-Carrasco, B.~Zhao, C.~Serrano, B.~Acha, T.~Serrano-Gotarredona,
  S.~Chen, and B.~Linares-Barranco, ``Mapping from frame-driven to frame-free
  event-driven vision systems by low-rate rate coding and coincidence
  processing--application to feedforward convnets,'' \emph{Pattern Analysis and
  Machine Intelligence, IEEE Transactions on}, vol.~35, no.~11, pp. 2706--2719,
  2013.

\bibitem{CAVIAR}
R.~Serrano-Gotarredona, M.~Oster, P.~Lichtsteiner, A.~Linares-Barranco,
  R.~Paz-Vicente, F.~G{\'o}mez-Rodr{\'\i}guez, L.~Camu{\~n}as-Mesa, R.~Berner,
  M.~Rivas-P{\'e}rez, T.~Delbruck \emph{et~al.}, ``{CAVIAR}: A 45k neuron, 5m
  synapse, 12g connects/s aer hardware sensory--processing--learning--actuating
  system for high-speed visual object recognition and tracking,'' \emph{IEEE
  Trans. Neural Netw.}, vol.~20, no.~9, pp. 1417--1438, 2009.

\bibitem{ShouShunAER}
S.~Chen, P.~Akselrod, B.~Zhao, J.~Perez-Carrasco, B.~Linares-Barranco, and
  E.~Culurciello, ``Efficient feedforward categorization of objects and human
  postures with address-event image sensors,'' \emph{Pattern Analysis and
  Machine Intelligence, IEEE Transactions on}, vol.~34, no.~2, pp. 302--314,
  2012.

\bibitem{reviewNeuromorphicSensorySystems}
S.-C. Liu and T.~Delbruck, ``Neuromorphic sensory systems,'' \emph{Current
  Opinion in Neurobiology}, vol.~20, no.~3, pp. 288 -- 295, 2010.

\bibitem{HMAX2007}
T.~Serre, L.~Wolf, S.~Bileschi, M.~Riesenhuber, and T.~Poggio, ``Robust object
  recognition with cortex-like mechanisms,'' \emph{Pattern Analysis and Machine
  Intelligence, IEEE Transactions on}, vol.~29, no.~3, pp. 411--426, 2007.

\bibitem{Sciences2006}
T.~Serre, ``{Learning a dictionary of shape-components in visual cortex:
  comparison with neurons, humans and machines},'' Ph.D. dissertation,
  Massachusetts Institute of Technology, 2006.

\bibitem{FOPEHMAX}
F.~Folowosele, R.~J. Vogelstein, and R.~Etienne-Cummings, ``Towards a cortical
  prosthesis: Implementing a spike-based hmax model of visual object
  recognition in silico,'' \emph{Emerging and Selected Topics in Circuits and
  Systems, IEEE Journal on}, vol.~1, no.~4, pp. 516--525, 2011.

\bibitem{shoushun2009bio}
C.~Shoushun, B.~Martini, and E.~Culurciello, ``A bio-inspired eventbased size
  and position invariant human posture recognition algorithm,'' \emph{IEEE Int.
  Symp. Circuits and Systems}, pp. 775--778, 2009.

\bibitem{masquelier2007unsupervised}
T.~Masquelier and S.~J. Thorpe, ``Unsupervised learning of visual features
  through spike timing dependent plasticity,'' \emph{PLoS Computational
  Biology}, vol.~3, no.~2, p.~31, 2007.

\bibitem{izhikevich2004model}
E.~M. Izhikevich, ``Which model to use for cortical spiking neurons?''
  \emph{IEEE Trans. Neural Netw.}, vol.~15, no.~5, pp. 1063--1070, 2004.

\bibitem{BestArchitectureForRecognition}
K.~Jarrett, K.~Kavukcuoglu, M.~Ranzato, and Y.~LeCun, ``What is the best
  multi-stage architecture for object recognition?'' \emph{Proc. 12th IEEE Int.
  Conf. Computer Vision}, pp. 2146--2153, Oct 2009.

\bibitem{Nielsen2015}
M.~A. Nielsen, \emph{{Neural Networks and Deep Learning}}.\hskip 1em plus 0.5em
  minus 0.4em\relax Determination Press, 2015.

\bibitem{Riesenhuber:1999}
M.~Riesenhuber and T.~Poggio, ``Hierarchical models of object recognition in
  cortex,'' \emph{Nat. Neurosci.}, vol.~2, no.~11, pp. 1019--1025, 1999.

\bibitem{FeedforwardRapidCategorization}
T.~Serre, A.~Oliva, and T.~Poggio, ``A feedforward architecture accounts for
  rapid categorization,'' \emph{Proc. National Academy of Sciences}, vol. 104,
  no.~15, pp. 6424--6429, 2007.

\bibitem{Orchard2013}
G.~Orchard, J.~G. Martin, R.~J. Vogelstein, and R.~Etienne-Cummings, ``{Fast
  neuromimetic object recognition using FPGA outperforms GPU
  implementations},'' \emph{IEEE Trans. Neural Networks Learn. Syst.}, vol.~24,
  no.~8, pp. 1239--1252, 2013.

\end{thebibliography}
\begin{IEEEbiography}[{\includegraphics[width=1in,height=1.25in,clip,keepaspectratio]{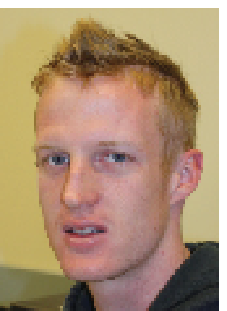}}]{Garrick Orchard}
received the B.Sc. degree in electrical engineering from the University of Cape Town, South Africa (2006) and the M.S.E. and Ph.D. degrees in electrical and computer engineering from Johns Hopkins University, Baltimore (2009, 2012). He has been named a Paul V. Renoff fellow (2007) and a Virginia and Edward M. Wysocki, Sr. fellow (2011). He has received the JHUAPL Hart Prize for Best R\&D Project (2009), and the IEEE BioCAS best live demo award (2012). He is currently a Postdoctoral Research Fellow at the Singapore Institute for Neurotechnology (SINAPSE) at the National University of Singapore where his research focuses on developing neuromorphic vision sensors and algorithms for high speed vision tasks. His other research interests include mixed-signal VLSI, compressive sensing, navigation, and legged locomotion.
\end{IEEEbiography}

\begin{IEEEbiography}[{\includegraphics[width=1in,height=1.25in,clip,keepaspectratio]{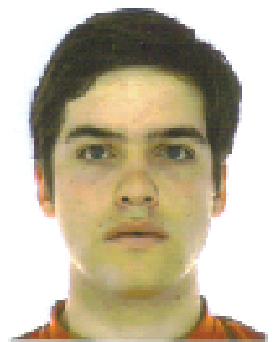}}]{Cedric Meyer}
received the B.Sc. degree in electrical engineering from the Ecole Nationale Supérieure de Cachan, France in 2010 and a ph.D degree in robotics from University Pierre and Marie Curie, Paris, France in 2013. His current research interests include efficient vision perception and computation for mobile robotics. He is also interested in understanding how biological visual systems encode and process visual information to perform object recognition.
\end{IEEEbiography}

\begin{IEEEbiography}[{\includegraphics[width=1in,height=1.25in,clip,keepaspectratio]{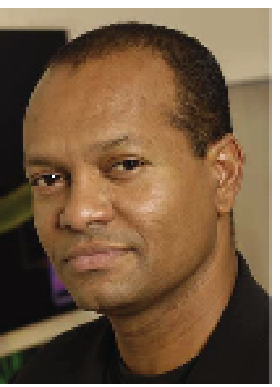}}]{Ralph Etienne-Cummings}
received his B. Sc. in physics, 1988, from Lincoln University, Pennsylvania. He completed his M.S.E.E. and Ph.D. in electrical engineering at the University of Pennsylvania in 1991 and 1994, respectively. He is currently a professor of electrical and computer engineering, and computer science at Johns Hopkins University (JHU). He is the former Director of Computer Engineering at JHU and the Institute of Neuromorphic Engineering. He is also the Associate Director for Education and Outreach of the National Science Foundation (NSF) sponsored Engineering Research Centers on Computer Integrated Surgical Systems and Technology at JHU. He has served as Chairman of the IEEE Circuits and Systems (CAS) Technical Committee on Sensory Systems and on Neural Systems and Application. He was also the General Chair of the IEEE BioCAS 2008 Conference. He was also a member of Imagers, MEMS, Medical and Displays Technical Committee of the ISSCC Conference from 1999 – 2006. He is the recipient of the NSF's Career and Office of Naval Research Young Investigator Program Awards. In 2006, he was named a Visiting African Fellow and a Fulbright Fellowship Grantee for his sabbatical at University of Cape Town, South Africa. He was invited to be a lecturer at the National Academies of Science Kavli Frontiers Program, in 2007. He has won publication awards including the 2003 Best Paper Award of the EURASIP Journal of Applied Signal Processing and "Best Ph.D. in a Nutshell" at the IEEE BioCAS 2008 Conference, and has been recognized for his activities in promoting the participation of women and minorities in science, technology, engineering and mathematics. His research interest includes mixed signal VLSI systems, computational sensors, computer vision, neuromorphic engineering, smart structures, mobile robotics, legged locomotion and neuroprosthetic devices.
\end{IEEEbiography}

\begin{IEEEbiography}[{\includegraphics[width=1in,height=1.25in,clip,keepaspectratio]{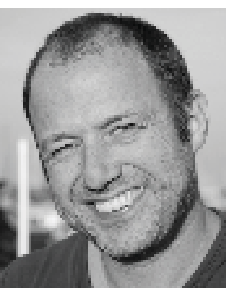}}]{Christoph Posch}
received the M.Sc. and Ph.D. degrees in electrical engineering and experimental physics from Vienna University of Technology, Vienna, Austria, in 1995 and 1999, respectively. From 1996 to 1999, he worked on analog CMOS and BiCMOS IC design for particle detector readout and control at CERN, the European Laboratory for Particle Physics in Geneva, Switzerland. From 1999 onwards he was with Boston University, Boston, MA, engaging in applied research and analog/mixed-signal integrated circuit design for high-energy physics instrumentation. In 2004 he joined the newly founded Neuroinformatics and Smart Sensors Group at AIT Austrian Institute of Technology (formerly Austrian Research Centers ARC) in Vienna, Austria, where he was promoted to Principal Scientist in 2007. Since 2012, he is co-directing the Neuromorphic Vision and Natural Computation group at the Institut de la Vision in Paris, France, and has been appointed Associate Research Professor at Universite Pierre et Marie Curie, Paris 6. His current research interests include biomedical electronics, neuromorphic analog VLSI, CMOS image and vision sensors, and biology-inspired signal processing. Dr. Posch has been recipient and co-recipient of several scientific awards including the Jan van Vessem Award for Outstanding European Paper at the IEEE International Solid-State Circuits Conference (ISSCC) in 2006, the Best Paper Award at ICECS 2007, and Best Live Demonstration Awards at ISCAS 2010 and BioCAS 2011. He is senior member of the IEEE and member of the Sensory Systems and the Neural Systems and Applications Technical Committees of the IEEE Circuits and Systems Society. Christoph Posch has authored more than 80 scientific publications and holds several patents in the area of artificial vision and image sensing.
\end{IEEEbiography}

\begin{IEEEbiography}[{\includegraphics[width=1in,height=1.25in,clip,keepaspectratio]{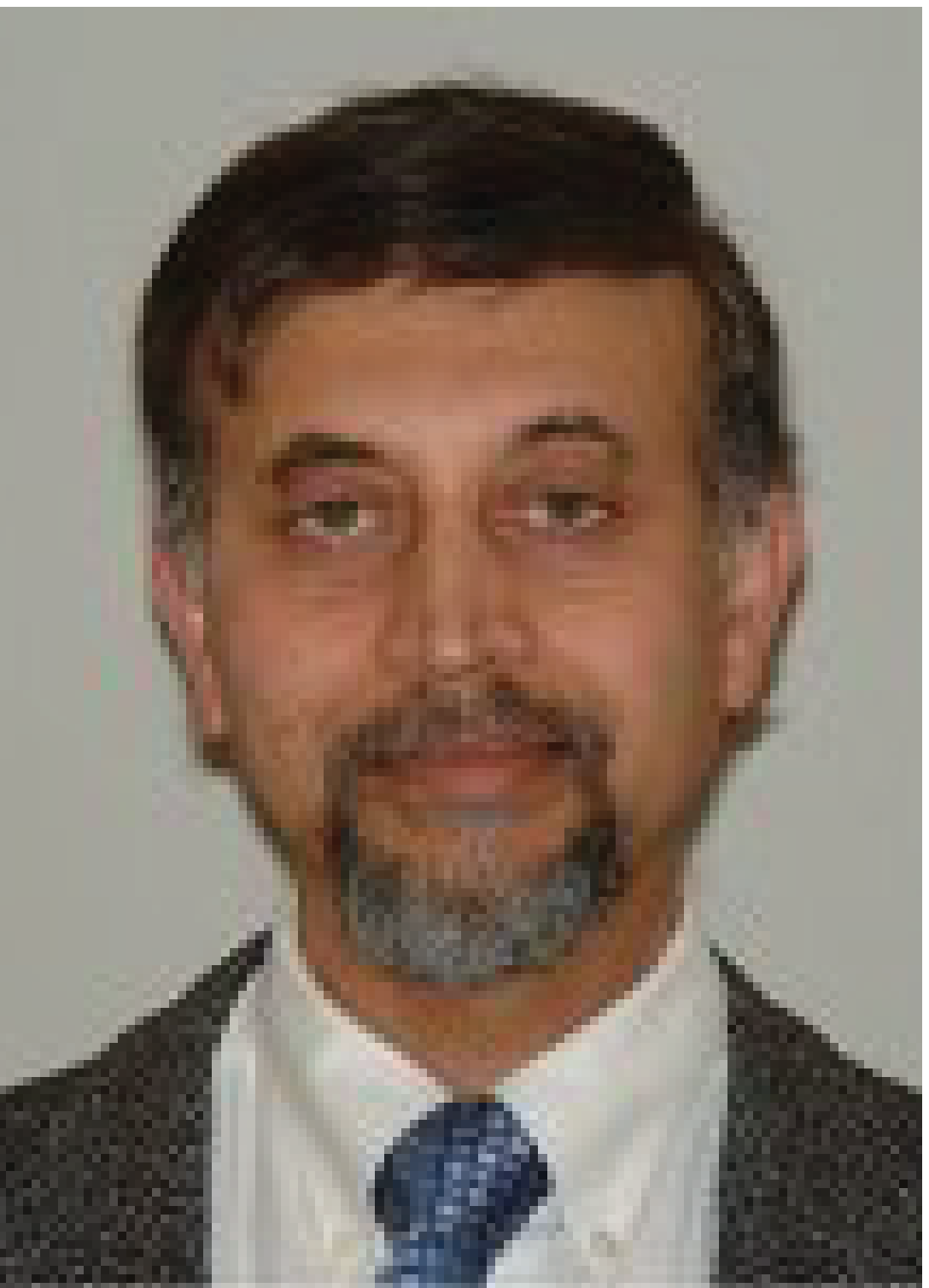}}]{Nitish Thakor}
(S'78–M'81–SM'89–F'97) is a Professor of  Biomedical Engineering, Electrical and Computer Engineering, and Neurology at Johns Hopkins University, Baltimore, MD, and directs the Laboratory for Neuroengineering. He has been appointed as the Provost Professor, National University of Singapore, and now leads the SiNAPSE Institute, focused on neurotechnology research and development.His technical expertise is in the areas of neural diagnostic instrumentation, neural 
microsystem, neural signal processing, optical imaging of the nervous system, rehabilitation, neural control of prosthesis and brain machine interface. He is the Director of a Neuroengineering Training program funded by the National Institute of Health. He has authored 250 refereed journal papers, generated 11 patents, cofounded four companies, and carries out research funded mainly by the NIH, NSF and DARPA. He was the Editor-in-Chief of IEEE Transactions On Neural And Rehabilitation Engineering (2005–2011) and is currently the Editor-in-Chief of Medical and Biological Engineering and Computing journal. Dr. Thakor is a recipient of a Research Career Development Award from the National Institutes of Health and a Presidential Young Investigator Award from the National Science Foundation. He is a Fellow of IEEE, the American Institute of Medical and Biological Engineering, International Federation of Medical and Biological Engineering, and Founding Fellow of the Biomedical Engineering Society, Technical Achievement Award from IEEE and Distinguished Alumnus award from Indian Institute of Technology, Bombay and University of Wisconsin, Madison.
\end{IEEEbiography}

\begin{IEEEbiography}[{\includegraphics[width=1in,height=1.25in,clip,keepaspectratio]{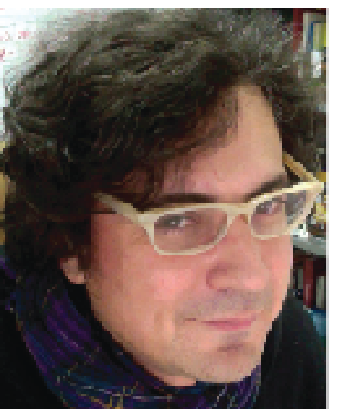}}]{Ryad Benosman}
received the M.Sc. and Ph.D. degrees in applied mathematics and robotics from University Pierre and Marie Curie in 1994 and 1999, respectively. He is Associate Professor with University Pierre and Marie Curie, Paris, France, leading the Natural Computation and Neuromorphic Vision Laboratory, Vision Institute, Paris. His work covers neuromorphic visual computation and sensing. He is currently involved in the French retina prosthetics project and in the development of retina implants and cofounder of Pixium Vision a french prosthetics company. He is an expert in complex perception systems, which embraces the conception, design, and use of different vision sensors covering omnidirectional 360 degree wide-field of view cameras, variant scale sensors, and non-central sensors. He is among the pioneers of the domain of omni- directional vision and unusual cameras and still active in this domain. He has been involved in several national and European robotics projects, mainly in the design of artifcial visual loops and sensors. His current research interests include the understanding of the computation operated along the visual systems areas and establishing a link between computational and biological vision. Ryad Benosman has authored more than 100 scientific publications and holds several patents in the area of vision, robotics and image sensing. In 2013 he was awarded with the national best French scientific paper by the publication La Recherche for his work on neuromorphic retinas.
\end{IEEEbiography}

\end{document}